\documentclass[twocolumn,showpacs,preprintnumbers,amsmath,amssymb]{revtex4}

\usepackage{graphicx}
\usepackage{dcolumn}
\usepackage{bm}

\usepackage{subfigure}
\RequirePackage{epsfig}
\RequirePackage{subeqn}
\RequirePackage{RandZ}
\RequirePackage{theorem}

\begin{document}

\preprint{APS/123-QED}

\title{Cauchy Annealing Schedule: \\ An Annealing Schedule for
Boltzmann Selection Scheme in 
Evolutionary Algorithms}

\author{Ambedkar Dukkipati}
 \email{ambedkar@csa.iisc.ernet.in}
\author{M. Narasimha Murty}
 \email{mnm@csa.iisc.ernet.in} 
\author{Shalabh Bhatnagar}
 \email{shalabh@csa.iisc.ernet.in}
\affiliation{
Department of Computer Science and Automation
Indian Institute of Science, Bangalore 560012, India.
}

\date{\today}

\begin{abstract}
	Boltzmann selection is an important selection mechanism
	in evolutionary algorithms as it has theoretical
	properties which help in theoretical analysis. However, Boltzmann
	selection is not used in practice 
	because a good annealing schedule for the `inverse
	temperature' parameter is lacking. In this paper 
	we propose a Cauchy annealing schedule for Boltzmann selection scheme
	based on a hypothesis that selection-strength should increase
	as evolutionary process goes on and distance between two selection
	strengths should decrease for the process to
	converge. 
	To formalize these aspects, we develop formalism for
	selection mechanisms using fitness distributions and give an
	appropriate measure for selection-strength.
	In this paper, we prove an important result, by which we 
	derive an annealing schedule 
	called Cauchy annealing schedule. We demonstrate the novelty
	of proposed annealing schedule using simulations in the framework of 
	genetic algorithms.
\end{abstract}

\maketitle

\section{Introduction}
\label{Section:Introduction}

	Selection is a central concept in evolutionary algorithms. 
	There are several selection mechanisms in genetic
	algorithms, like proportionate selection, ranking selection,
	tournament selection, truncation selection and Boltzmann
	selection~\cite{BlickleThiele:1996:ComparisionOfSelectionSchems}.
	Among all these
	selection mechanisms, Boltzmann selection has an  
	important place because it has some nice theoretical 
	properties in some models of evolutionary
	algorithms~\cite{MahnigMuhlenbein:2001:AdaptiveBoltzmannSelection}.
	For example, Boltzmann selection is extensively used in
	statistical 
	mechanics theory of evolutionary algorithms
	~\cite{PrugelShapiro:1994:GAStatisticalMechanics,Prugel:1997:ModellingEvolvingPopulations,PrugelRogers:2001:ModellingGAdynamics,Rattray:1995:DynamicsOfGAunderStabilizingSelection}.

	Moreover, Boltzmann selection scheme is not used often in solving
	practical problems because,
	similar to simulated annealing, it needs an annealing schedule for
	perturbing the value of inverse temperature 
	parameter used in Boltzmann selection, which is difficult to
	choose~\cite{MahnigMuhlenbein:2001:AdaptiveBoltzmannSelection}.
	This problem is well known from simulated
	annealing~\cite{AartsKorst:1989:SimulatedAnnealingAndBoltzmannMachines},
	an optimization algorithm where noise is introduced by means
	of a formal temperature. Lowering, or ``annealing,'' the 
	temperature from high to low values in the course of the
	optimization leads to improved results compared to an
	optimization at fixed
	temperature~\cite{Bornholdt:1999:AnnelingScheduleFromPopulationDynamics}.
	However, there remains the problem of choosing a suitable
	annealing schedule for a given optimization problem. The same
	problem occurs in population-based optimization algorithms,
	and this paper address this problem for evolutionary algorithms.

	Usually, in evolutionary algorithms, probabilistic selection
	mechanisms are characterized by {\em selection
	probabilities}~\cite{Back:1994:SelectivePressure}. For a population
	$P = \{\omega_{i}\}_{i=1}^{n_{P}}$, selection probabilities
	$ \{ p_{i} \}_{i=1}^{n_{P}}$ are defined as,
	\[  p_{i} = \mbox{Prob}( \omega_{i} \in \mbox{selection}(P)
	| \omega_{i}\in P)  \: \: \forall i =1 \ldots n_{P}\enspace, 
	\]
	\noindent and $ \{ p_{i} \}_{i=1}^{n_{P}}$ satisfies the
	condition: $\sum_{i=1}^{n_{P}} p_{i} = 1. $ 

	Let $\{f(\omega_{i})\}_{i=1}^{n_{P}}$ be the corresponding fitness
	values. The proportionate selection assigns selection
	probabilities according to the relative fitness of
	individuals as~\cite{Holland:1975:Adaptation}: 
	\begin{equation}
            p_{i} = \frac{f(\omega_{i})}{\sum_{j=1}^{n_{P}} f(\omega_{j})}\enspace.
	\end{equation}
	\noindent Similarly Boltzmann selection is represented
	as~\cite{MazaTidor:1993:ProportionalBoltzmannSelection}:
	\begin{equation}
            p_{i} = \frac{e^{\gamma f(\omega_{i})}}{\sum_{j=1}^{n_{P}} e^{\gamma
            f(\omega_{j})}}\enspace,
	\end{equation}

	\noindent where $\gamma$ is called {\em inverse temperature}. The strength of
        selection is controlled by the parameter $\gamma$. A higher
        value of $\gamma$ (low temperature) gives a stronger
        selection, and a lower value of $\gamma$ gives a weaker
        selection. 
	For details of representation of other selection mechanisms
        refer~\cite{Back:1994:SelectivePressure,BlickleThiele:1996:ComparisionOfSelectionSchems,WieczorekCzech:2002:SelectionSchemsInEA}.

	Some properties of selection mechanisms that are 
	desirable in order to control the search
	process are~\cite{Back:1994:SelectivePressure}: 
	\begin{itemize}
	  \item The impact of the control parameters on selective
	  pressure should be simple and predictable.

	  \item One single control parameter for selective pressure is
	  preferable.

	  \item The range of selective pressure that can be realized by
	  varying the control parameter should be as large as
	  possible.
	\end{itemize}
	Boltzmann selection satisfies above
	properties. Boltzmann
	selection gives faster convergence, but without good
	annealing schedule for $\gamma$, it might lead to premature
	convergence.

	In this paper we propose Cauchy criteria for choosing the
	Boltzmann 
	selection schedule. Based on this we derive an annealing
	schedule for the inverse temperature parameter $\gamma$, using
	a result we proved.
	Since selection depends only on the fitnesses of candidate
	solutions of population,
	in this paper we characterize the selection using {\em normalized
	fitness distribution} (normalized fitness distribution is
	precisely normalization of fitness distribution of
	population) instead of selection probabilities which are
	defined for all the members of population.  We also give a new
	measure for selection-strength which is suitable for the
	theoretical analysis presented in this
	paper.

	The outline of the paper is as follows. In \S~\ref{Section:SelectionSchemes}, 
	we present the formalization   
	of selection methods. We present our main results regarding Cauchy
	criteria for Boltzmann selection schedule in
	\S~\ref{Section:CauchyCriteria}. We present simulation
	results in \S~\ref{Section:SimulationResults}.

\section{A Formalization of Selection Schemes}
\label{Section:SelectionSchemes}

  \subsection{Definitions}
  \label{SubSection:SelectionSchems:Definitions} 
	Let $f:\Omega \rightarrow \bbbr^{+} \cup \{0\}$ be a fitness function, where
	$\Omega$ is the search space. Let $P =
	\{\omega_{k}\}_{k=1}^{n_{P}}$ denote the population. Here we assume
	that the size of population at any time is finite and need not be
	a constant.

	Fitness distribution is an important macroscopic property of
	population. Formal definition of fitness distribution of a
	population is given below~\cite{Dukkipati:2003:CEC2003}.

	\begin{definition}
	\label{Definition:FitnessDistributionOfPopulation}
         Fitness distribution of a population $P = \{ \omega_{k}  
         \}_{k=1}^{n_{P}}$ is a function $\rho^{P}: \bbbr
	\rightarrow {\bbbz}^{+}\cup \{0\}$ defined as
	\begin{equation}
	\rho^{P}(x) = \sum_{k=1}^{n_{P}} \delta (x - f(\omega_{k}))\enspace,  
	\end{equation}
	where $\delta:\bbbr \rightarrow \{0,1\}$ is the Kronecker
	delta function defined as $\delta(x)=1$ if $x=0$,
	$\delta(x)=0$ otherwise . 
	\end{definition}

	\noindent $\rho^{P}$ assigns each $x \in
	\bbbr$, the number of individuals in a population $P$
	carrying $x$ as the fitness value.
	The finite set of values associated with the fitness
	distribution which are mapped to non-zero values is called
	{\em support} of fitness distribution of population.
	\begin{definition}
	\label{Definition:FitnessValuesSet}
	Let $\rho^{P}$ be the fitness distribution of population $P$,
	then `support' of $\rho^{P}$ is defined
	as~\footnote{The actual definition of support of $\rho^{P}$ is
	$\overline{\{ x : \rho^{P}(x) \neq 0\}}$. The overline denotes
	the closure of the set. Since $\{ x : \rho^{P}(x) \neq 0\}$ is
	finite $\{ x : \rho^{P}(x) \neq 0\} = \overline{\{ x :
	\rho^{P}(x) \neq 0\}}$} 
	\begin{equation}
	  supp(\rho^{P}) = E_{\rho} \mbox{(or}E_{P}\mbox{)}
	= \{ x : \rho^{P}(x) \neq 0\} \enspace.
	  \end{equation}
	\end{definition}
	For any population $P$, $supp(\rho^{P})$ is finite set,
	since population size is finite.
	We can write size of a population $P$ in terms of
	its fitness
	distribution $\rho^{P}$ as, 
	\begin{equation}
	\label{Equation:SizeOfPopulation}
		n_{P} = \sum_{x \in E_{P}} \rho^{P}(x)\enspace. 
	\end{equation}

	We now define {\em normalized fitness distribution} (NFD).
	\begin{definition}
	\label{Definition:nfdOfPopulation}
         Normalized fitness distribution (NFD) of a population $P =
	\{ \omega_{k} \}_{k=1}^{n_{P}}$ with fitness distribution
	$\rho^{P}$ is a function $\varphi^{P}: \bbbr \rightarrow [0,1]$
	defined as 
		\begin{equation}
		\varphi^{P}(x) = \frac{\rho^{P}(x)}{n_{P}}\enspace, \:\:\: \forall x \in \bbbr \enspace.
		\end{equation}
	\end{definition}

	\noindent One can see that $\varphi^{P}$ is well
	defined. From~(\ref{Equation:SizeOfPopulation}), we have
	\begin{equation}
	\sum_{x \in E_{P}}^{} \varphi^{P}(x)=1 \enspace.
	\end{equation}

	\noindent Note that $supp(\varphi^{P}) =
	supp(\rho^{P})$. Support of a NFD $\varphi$ of population $P$ is
	represented by 
	$E_{\varphi}$. 

  \subsection{Representation of Selection Schemes Via NFD}
  \label{SubSection:SelectionSchems:RepresentationUsingNFD} 
	Instead of giving a mechanistic view of selection, we define
	selection as an operator on fitness distribution (hence on
	NFD). For that we need to specify the corresponding space.

	Definition~\ref{Definition:nfdOfPopulation} gives the
	definition of ``NFD of a population''. To define space of
	all NFDs we give a generalized definition of NFD, similar to
	the generalized definition of fitness distribution given
	in~\cite{Dukkipati:2003:CEC2003}.
	\begin{definition}
	\label{Definition:nfd}
	  `Normalized fitness distribution' (NFD) is a function $\varphi :\bbbr
          \rightarrow [0,1]$ which satisfies
	  \begin{subequations}
	    \begin{equation}
		   \sharp \{ x : \varphi(x) \neq 0\}  < \infty \:\:\:
	(i.e., \sharp supp(\varphi) < \infty ) \enspace,
	    \end{equation}
	    \begin{equation}
                   \sum_{x\in supp(\varphi)} \varphi(x) = 1 \enspace,
	    \end{equation}
	  \end{subequations}
		where $\sharp$ denotes the cardinality of a set.
	\end{definition}

	\noindent From Definition~\ref{Definition:nfdOfPopulation},
	one can easily see that every 
	``NFD of a population'' is indeed an ``NFD''. Space of all
	NFDs is denoted by $\mathcal{O}$ i.e.,
	\begin{equation}
	\mathcal{O} =  \{ \varphi:\bbbr \rightarrow [0,1] \::\: \sharp
	supp(\varphi) < \infty, \sum_{x\in supp(\varphi)} 
	\varphi(x) = 1 \} \enspace.
	\end{equation}
	\noindent We define selection as an operator $\Gamma$ on the space
	$\mathcal{O}$ i.e., $\Gamma: \mathcal{O} \rightarrow
	\mathcal{O}$. At generation $k$, for a population $P_{k}$, with fitness
	distribution $\rho^{k}$ and population size $N_{k}$,
	Boltzmann selection $\Gamma$ can be represented in terms of
	fitness distribution as
	\begin{equation}
	\rho^{k+1}(x) = \Gamma \rho^{k}(x) = \rho^{k}(x) \frac{e^{\gamma
	x}}{\sum_{y\in 
	E} \rho^{k}(y)e^{\gamma y}} N_{k+1} \enspace,
	\end{equation}
	\noindent where $N_{k+1}$ is the population size after the
	selection $\Gamma$ and $E = supp(\rho^{k})$. From
	Definition~\ref{Definition:nfdOfPopulation}, we have
	\begin{eqnarray}
	 \varphi^{k+1}(x) = \frac{\Gamma \rho^{k}(x)}{N_{k+1}} = 
	\rho^{k}(x) \frac{e^{\gamma x}}{\sum_{y\in E}^{}
	 \rho^{k}(y)e^{\gamma y}}  \nonumber \\ =
	 \frac{\varphi^{k}(x)}{N_{k}} \frac{e^{\gamma x}}{\sum_{y\in E}^{} 
	 \frac{\varphi^{k}(y)}{N_{k}} e^{\gamma y}} \enspace. \nonumber
	 \end{eqnarray}
	\[
         \varphi^{k+1}(x) =  \frac{\varphi^{k}(x) e^{\gamma x}}{\sum_{y\in E}^{} 
         \varphi^{k}(y) e^{\gamma y}} \enspace.   
	\]
	\noindent Hence Boltzmann selection operator $\Gamma$ on $\mathcal{O}$ is
	defined as 
	\begin{equation}
	\label{Equation:DefinitionOfBoltzmannSelection}
	\Gamma \varphi(x) = \frac{\varphi(x) e^{\gamma x}}{\sum_{y\in E}^{} 
         \varphi(y)e^{ \gamma y}} \enspace,\:\:\: \forall x \in \bbbr \enspace, \:\:\:
	\forall \varphi \in \mathcal{O} \enspace,
	\end{equation}
	where $\gamma \in \bbbr^{+} \cup \{0\}$ corresponds to inverse
	temperature. Similarly we can define proportionate selection using operator 
	$\Gamma_{\mbox{prop}}$ as follows: 
	\begin{equation}
	\label{Equation:DefinitionOfProportionateSelection}
	\Gamma_{\mbox{prop}}\varphi(x) = \frac{x\varphi(x)}{\sum_{y\in E}^{} y
         \varphi(y)} \enspace,\:\:\: \forall x \in \bbbr \enspace,\:\:\:
	\forall \varphi \in \mathcal{O} \enspace.
	\end{equation}
	Through out this paper we represent Boltzmann selection by
	$\Gamma$ unless mentioned otherwise.


  \subsection{Metric on Space of NFDs}
  \label{SubSection:SelectionSchems:Metric} 
	One can view NFD as a probability distribution and one can
	use various distance measures on it. For example, one can use
        Kullback-Leibler distance measure but it is not a
        metric~\cite{KesavanKapur:1997:EntropyOptimizationPrinciples}. We  
	define a metric $d:
	\mathcal{O} \times \mathcal{O} \rightarrow \bbbr $ according to 
	\begin{equation}
	\label{Equation:MetricOnNfd}
	d(\varphi_{1}, \varphi_{2}) = \sum_{x\in E_{\varphi_{1}} \cup
  E_{\varphi_{2}}}^{} \left| 
\varphi_{1}(x) - \varphi_{2}(x) \right| \enspace,\:\:\:\: \forall \varphi_{1}, \varphi_{2} 
	\in \mathcal{O} \enspace. 
	\end{equation}
	\noindent It is easy to verity that $d$ is indeed a
	metric on  $\mathcal{O}$.

  \subsection{Selection Strength}
  \label{SubSection:SelectionSchems:SelectionStrength} 
	
	There have been several variants to measure selection strength in
	evolutionary algorithms. The terminology  ``selection
	intensity'' or ``selection pressure'' is often used to
	describe this property of selection.

	The concept of ``take
	over time'' quantifies selection pressure by the number of
	generations required by repeated application of selection, to
	fill the complete population with copies of the single
	initially best
	individual~\cite{GoldbergDeb:1991:ComparisionOfSelectionSchemes}.
	There have been some adaptations of definitions from population
	genetics for selection
	intensity. The
	change in average fitness of the population due to
	selection is a reasonable measure of selection
	intensity~\cite{Muhlenbein:1993:PredictiveModelsForBGA}. Also
	note that several of these measures depend on fitness
	distribution at that instance. Details of selection intensity
	measures can be found
	in~\cite{GoldbergDeb:1991:ComparisionOfSelectionSchemes,Muhlenbein:1993:PredictiveModelsForBGA,Back:1994:SelectivePressure}.




	We measure selection strength w.r.t an NFD
	using the metric $d$ as distance between the NFD before the
	selection and after selection. Let $\Gamma: \mathcal{O} \rightarrow
	\mathcal{O}$ be the selection operator. The selection strength
	can be measured as:
	\begin{equation}
	d(\varphi,\Gamma \varphi) = \sum_{x \in E_{\varphi} }^{} \left|
	\varphi(x) - \Gamma \varphi(x) \right| \enspace.
	\end{equation}

	We give the formal definition of selection strength as
	follows.
	\begin{definition}
	\label{Definition:SelectionStrength}
	Selection strength of a selection scheme $\Gamma$ with respect
	to an NFD $\varphi \in \mathcal{O}$ is denoted by
	$S_{\varphi}(\Gamma)$ and is defined as
	\begin{equation}
	   S_{\varphi}(\Gamma) = d(\varphi, \Gamma\varphi) \enspace.
	\end{equation}
	\end{definition}
	For example, for proportionate selection the NFD $\varphi$ selection
	strength can be measured as:

	\begin{eqnarray}
	d(\varphi,\Gamma_{prop} \varphi) = \sum_{x \in
	E_{\varphi}}^{} \left|\varphi(x) - 
	\frac{x\varphi(x)}{\sum_{y \in E_{\varphi}}^{} y\varphi(y) }
	\right| \nonumber \\ 
	= \sum_{x \in E_{\varphi}} \varphi(x) \left| \frac{\sum_{y \in
	E_{\varphi}}^{} y\varphi(y) 
	- x} {\sum_{y \in E_{\varphi}}^{} y\varphi(y)}\right| \enspace, 
	\end{eqnarray}

	\begin{equation}
	\label{Equation:SelectionStrenghtForProportionateSelection}
	d(\varphi, \Gamma_{prop} \varphi) = \frac{\sum_{x \in E_{\varphi}}
	\varphi(x) \left| \mu_{\varphi} -x 
	\right| }{\mu_{\varphi}} \enspace.
	\end{equation}

	\noindent where $\mu_{\varphi} = \sum_{x \in E_{\varphi}} x
	\varphi(x)$ is expectation of $\varphi$. The numerator 
	is nothing but mean absolute error of $\varphi$. If one
	observes~(\ref{Equation:SelectionStrenghtForProportionateSelection}) 
	carefully,
	it justifies the definition of selection strength as
	$d(\varphi,\Gamma \varphi)$.

\section{Cauchy Criteria for Boltzmann Selection Scheme}
\label{Section:CauchyCriteria}

  \subsection{Boltzmann Selection Scheme}

	Let $\{P_{n}\}$ be the evolutionary process, where $P_{n}$ is
	population at generation $n$. We represent corresponding
	Boltzmann selection scheme as
	$\{\Gamma_{(n)}\}$ where $\Gamma_{(n)}$ is an operator
	$\Gamma_{(n)}:\mathcal{O} \rightarrow \mathcal{O}$ and is defined as:
	\begin{eqnarray}
	\varphi_{n}(x) = \Gamma_{(n)}\varphi_{n-1}(x)  = \frac{\varphi_{n-1}(x)
	e^{\gamma_{n}x}}{\sum_{y\in E_{\varphi_{n-1}}}^{}  
	\varphi_{n-1}(y) e^{\gamma_{n}y}} \enspace, \nonumber \\
	\forall x \in \bbbr \enspace, 
	\:\:\: \forall n = 1,2, \ldots \enspace,
	\end{eqnarray}
	where $\varphi_{n} \in \mathcal{O}$. $\{\gamma_{n}\}$ is
	annealing schedule for the Boltzmann selection scheme
	$\{\Gamma_{(n)}\}$ and $\gamma_{n} \geq 0 \:\: \forall n=1,2
	\ldots $. Also $\{\gamma_{n}\}$ is a non-decreasing sequence
	since $\gamma_{n}$ represents the inverse
	temperature~\cite{MahnigMuhlenbein:2001:AdaptiveBoltzmannSelection}. 

  \subsection{Cauchy Criteria}

	Our Hypothesis for Boltzmann selection schedule is:
	\begin{quote}
	   The difference between successive selection pressures should
           decrease as the evolutionary process proceeds.
	\end{quote}

	We formalize above hypothesis as Cauchy criteria for Boltzmann
	selection schedule as follows:
	\begin{definition}
	A Boltzmann selection schedule $\{\Gamma_{(n)}\}$ is said to satisfy
	Cauchy criteria if
	$\{\Gamma_{(n)} \varphi\} \subset \mathcal{O}$ is Cauchy with
	respect to metric $d$, $\:\:\: \forall
	  \varphi \in \mathcal{O}$.
	\end{definition}
	\noindent We justify the fact that Cauchy criteria for
	Boltzmann selection schedule 
	captures the hypothesis by the following lemma.
	\begin{lemma}
	Let $\Gamma_{1}$ and $\Gamma_{2}$ be two Boltzmann selection
	operators. Then for any $\varphi \in \mathcal{O}$, difference
	between these selection strengths satisfies
	\begin{equation}
 	\left| S_{\varphi}(\Gamma_{1}) - S_{\varphi}(\Gamma_{2})
 	\right| \leq
	d(\Gamma_{1}\varphi, \Gamma_{2}\varphi) \enspace.
	\end{equation}
	\end{lemma}

 	\begin{proof}
 	From Definition~\ref{Definition:SelectionStrength} we have
 	\[
 	\left| S_{\varphi}(\Gamma_{1}) - S_{\varphi}(\Gamma_{2})
 	\right| = \left| d(\varphi, \Gamma_{1}\varphi) -
 	d(\varphi,\Gamma_{2} \varphi) \right| \enspace.
 	\]
 	From triangular inequality we have
 	\[
 	d(\varphi, \Gamma_{1}\varphi) \leq d(\varphi,
 	\Gamma_{2}\varphi) + d(\Gamma_{1} \varphi, \Gamma_{2} \varphi)
 	\enspace,
 	\]
	\noindent which gives
 	\begin{subequations}
 	\begin{equation}
 	\label{Equation:LemmaSelectionStrengthTriangular1}
 	d(\Gamma_{1} \varphi, \Gamma_{2} \varphi)
 	\geq
 	d(\varphi,\Gamma_{1}\varphi) - d(\varphi, \Gamma_{2}\varphi) \enspace.
 	\end{equation}
 	\noindent Similarly we have
 	\begin{equation}
 	\label{Equation:LemmaSelectionStrengthTriangular2}
 	d(\Gamma_{1} \varphi, \Gamma_{2} \varphi)
 	\geq
 	d(\varphi,\Gamma_{2}\varphi) - d(\varphi, \Gamma_{1}\varphi) \enspace.
 	\end{equation}
 	\end{subequations}
 	From~(\ref{Equation:LemmaSelectionStrengthTriangular1}) and
 	(\ref{Equation:LemmaSelectionStrengthTriangular2}) we get
 	\[
 	d(\Gamma_{1} \varphi, \Gamma_{2} \varphi)
 	\geq
 	\left| d(\varphi,\Gamma_{1}\varphi) - d(\varphi,
 	\Gamma_{2}\varphi) \right| \enspace.
 	\]

 	\end{proof}

	Hence decrement in $d(\Gamma_{1}\varphi,
	\Gamma_{2}\varphi)$ results in decrement in the difference
	between selection strengths. From the definition of Cauchy
	sequence justification is clear.

	Note that above criteria is stated in terms of the selection
	operator. Based on this we derive an annealing schedule for
	inverse temperature parameter $\gamma_{n}$ in the next
	section.

  \subsection{Derivation of Cauchy Annealing Schedule}

	We summarize Cauchy criteria for Boltzmann selection schedule
	$\{\Gamma_{(n)}\}$ 
	as:
	\begin{description}

	  \item[(CB1)] $\{\gamma_{n}\}$ is non-decreasing sequence

	  \item[(CB2)] $\{\Gamma_{(n)} \varphi\} \subset \mathcal{O}$ is
	  Cauchy $\forall \varphi \in \mathcal{O}$

	\end{description}
	\noindent For $\{\Gamma_{(n)}\}$ to satisfy (CB1) we define
	\begin{equation}
	\label{Equation:ChoiceOfGamma}
	   \gamma_{n} = \sum_{k=1}^{n} g_{k} \enspace, \:\:\: \mbox{where}
	   \: \{g_{k}\} \subset \bbbr^{+} \cup {0}, \:\:\: \forall n =
	1,2, \ldots \enspace.
	\end{equation}
	\noindent Clearly $\{\gamma_{n}\}$ is non decreasing sequence.
	Then Boltzmann selection schedule $\{\Gamma_{(n)}\}$ defined as
	\begin{equation}
	\label{Equation:BoltzmannSelectionOperator_Increasing}
	  \Gamma_{(n)} \varphi(x) = \frac{\varphi(x)\exp(x \sum_{k=1}^{n}
	g_{k})}{\sum_{y \in E_{\varphi}} 
	\varphi(y)\exp(y \sum_{k=1}^{n} g_{k})}\enspace, \:\:\:
	\forall x \in \bbbr\enspace, 
	\end{equation}
	for arbitrary $\{g_{k}\} \subset \bbbr^{+} \cup \{0\}$
	satisfies (CB1).
	Now we derive annealing schedule for $\left\{ \gamma_{n} =
	\sum_{k=1}^{n} g_{k} \right\}_{n} $ for the selection schedule
	$\{\Gamma_{(n)}\}$ to satisfy (CB2). First we prove following
	inequality.

	\begin{lemma}
	\label{Lemma:Inequality}
	Let $\{\Gamma_{(n)}\}$ be a sequence of Boltzmann selection
	operators defined as
	in~(\ref{Equation:BoltzmannSelectionOperator_Increasing}),
	then for any NFD $\varphi \in \mathcal{O}$, we have 
	\[
	d(\Gamma_{(n)}(\varphi), \Gamma_{(m)}(\varphi)) \leq 
	\sum_{x \in E_{\varphi}}^{} \left( \exp (x\sum_{k=m+1}^{n}g_{k} ) - 1
       \right)
	\]
	whenever $n > m$ and $n,m \in {\bbbz}^{+}$.
	\end{lemma}

	\begin{proof}
	Denote
	$$C_{n}(x) = \varphi(x) \exp(x \sum_{k=1}^{n}g_{k}) \:\:\:
\forall x \in E_{\varphi} \enspace.$$ 
	\noindent Then,
	\begin{eqnarray}
	\lefteqn{d(\Gamma_{(n)}(\varphi),
	\Gamma_{(m)}(\varphi)) = } \nonumber\\  
	 && \sum_{x\in E_{\Gamma_{(n)}(\varphi)} \cup E_{\Gamma_{(m)}(\varphi)}}^{}
	\left| \frac{C_{n}(x)}{\sum_{y\in E_{\varphi}}^{} C_{n}(y) } - 
	\frac{C_{m}(x)}{\sum_{y\in E_{\varphi}}^{} C_{m}(y) } \right|
\enspace. \nonumber 
	\end{eqnarray}
	\noindent Since $supp(\varphi)
	\supseteq supp(\Gamma_{n}(\varphi)) \cup
	supp(\Gamma_{m}(\varphi))$ and $supp(C_{n}) =
	supp(\varphi) \:\: \forall n $ we can write

	\begin{eqnarray}
	\lefteqn{d(\Gamma_{(n)}(\varphi),
	\Gamma_{(m)}(\varphi)) = } \nonumber \\  
	 & &\sum_{x\in E_{\varphi}}^{}
	\left| \frac{C_{n}(x)}{\sum_{y\in E_{\varphi}}^{} C_{n}(y) } - 
	\frac{C_{m}(x)}{\sum_{y\in E_{\varphi}}^{} C_{m}(y) } \right| \nonumber 
	\end{eqnarray}

	\begin{equation}
	\label{Equation:InLemma_2}
	\leq
        \left( \frac{1}{\sum_{x\in E_{\varphi}}^{} C_{m}(x)} \right)
	\left( \sum_{x\in E_{\varphi}}^{}
	\left| C_{n}(x) - C_{m}(x) \right| \right) \enspace,
	\end{equation}

	\noindent since for $n > m$, $C_{n}(x) \geq C_{m}(x), \:\:\:
	\forall x > 0$.

	 \noindent We have,
	\begin{eqnarray}
	C_{n}(x) &=& \varphi(x) \exp(x \sum_{k=1}^{n}g_{k}) \nonumber \\
	&=& \varphi(x) \exp(x \sum_{k=1}^{m}g_{k})
	\exp (x \sum_{k=m+1}^{n}g_{k}) \nonumber \\
	&=& C_{m}(x) \exp (x \sum_{k=m+1}^{n}g_{k}) \enspace, \:\: \forall x \in
	E_{\varphi} \enspace.
	\end{eqnarray}

	\noindent Hence we can write~(\ref{Equation:InLemma_2}) as

	\begin{eqnarray*}
	 \lefteqn{ d(\Gamma_{(n)}(\varphi),
	\Gamma_{(m)}(\varphi)) =   \left( \frac{1}{\sum_{x\in E_{\varphi}}^{} C_{m}(x)}
	 \right)  }  \nonumber \\
	 & & \left( \sum_{x\in E_{\varphi}}^{} \left| C_{m}(x)  \left( \exp(
	 x\sum_{k=m+1}^{n}g_{k}) - 1 \right)  \right| \right)  \nonumber 
	\end{eqnarray*}

	\begin{eqnarray*}
	\lefteqn {\leq \left( \frac{1}{ \sum_{x\in E_{\varphi}}^{}
	C_{m}(x)} \right) }\\ 
	& &\sqrt{\sum_{x\in E_{\varphi}}^{} \left\{ C_{m}(x)
	\right\}^{2} \sum_{x\in E_{\varphi}}^{} \left\{ \exp(  
	x\sum_{k=m+1}^{n}g_{k}) - 1  \right\}^{2}} \enspace,
	\end{eqnarray*}

	\noindent by Cauchy-Schwartz-Bunyakovsky inequality.

	\noindent Since $C_{m}(x)$ and $\exp( x\sum_{k=m+1}^{n}g_{k})
	- 1$ are positive, we have 
	
	\begin{eqnarray*}
	\lefteqn {d(\Gamma_{(n)}(\varphi),
	\Gamma_{(m)}(\varphi)) \leq  \left( \frac{1}{  \sum_{x\in E_{\varphi}}^{} C_{m}(x)}
	\right) } \\
	& &\sqrt{\left\{ \sum_{x\in E_{\varphi}}^{} {C_{m}(x)} \right\}^{2}
	\left\{ \sum_{x\in E_{\varphi}}^{} \exp(  
	x\sum_{k=m+1}^{n}g_{k}) - 1  \right\}^{2}} \enspace,
	\end{eqnarray*}

	\begin{equation}
      	d(\Gamma_{(n)}(\varphi),
	\Gamma_{(m)}(\varphi))   \leq  \sum_{x \in E_{\varphi}}^{} \left( \exp( x\sum_{k=m+1}^{n}g_{k})
	- 1 \right) \enspace.
	\end{equation}

	\end{proof}

	\noindent We now give our main result which gives condition on
	annealing schedule $\{\gamma_{n}\}$ for Boltzmann selection to
	satisfy Cauchy criteria.

	\begin{theorem}
	Let $\{\Gamma_{(n)}\}$ be a sequence of Boltzmann selection
	operators defined as
	in~(\ref{Equation:BoltzmannSelectionOperator_Increasing}).
	Then,
	\[
	  \left\{\sum_{k=1}^{n} g_{k} \right\}_{(n)} \:\: \mbox{is
	Cauchy} \Longrightarrow  \{\Gamma_{(n)}\varphi\}
	\:\:\: \mbox {is Cauchy}
	\]
	$\forall \varphi \in \mathcal{O}$ and  for any $\{g_{k}\} \subset
	\bbbr^{+} \cup \{0\}$.

	\end{theorem}

	\begin{proof}
	$\{\Gamma_{(n)} \varphi \}$ is Cauchy for any $\varphi \in
	\mathcal{O}$ if
	\[ \forall \epsilon > 0 , \exists N = N(\epsilon) \in
	{\dZ}^{+} \ni \]
	\[  n, m \geq N \Rightarrow d(\Gamma_{(n)}(\varphi),
	\Gamma_{(m)}(\varphi) ) < \epsilon \enspace.\]

	\noindent Now consider $d(\Gamma_{(n)}(\varphi), \Gamma_{(m)}(\varphi)) $.
	With out loss of generality assume that $n > m$.
	From Lemma~\ref{Lemma:Inequality} we have
	\[
	d(\Gamma_{(n)}(\varphi), \Gamma_{(m)}(\varphi)) \leq 
	\sum_{x \in E_{\varphi}}^{} \left( \exp (x\sum_{k=m+1}^{n}g_{k} ) - 1
       \right) \enspace.
	\]

	\noindent Let $\epsilon > 0 $ arbitrary. So,

       \begin{eqnarray}	
        \sum_{x \in E_{\varphi}}^{} \left( \exp (x\sum_{k=m+1}^{n}g_{k} )
        - 1 \right) < \epsilon \Longrightarrow \nonumber \\
        d(\Gamma_{(n)}(\varphi), \Gamma_{(m)}(\varphi)) < \epsilon \enspace.
	\end{eqnarray}

	\noindent Hence it is enough to prove that
	\begin{eqnarray}
	\exists N = N(\epsilon) \in
	{\bbbz}^{+} \ni n, m \geq N \Rightarrow \nonumber \\ \sum_{x \in
	E_{\varphi}}^{} \left( \exp (x\sum_{k=m+1}^{n}g_{k} ) - 1
	\right) < \epsilon  \enspace.
	\end{eqnarray}

	\noindent Now let $E_{\varphi} =\{x_{i}\}_{i=1}^{r}$. $r <
	\infty$ since $E_{\varphi}$ is 
	finite. We thus have to prove that

	\begin{eqnarray}
	 \exists N = N(\epsilon) \in {\dZ}^{+} \ni 
	n, m \geq N \Longrightarrow  \nonumber \\
	\sum_{i=1}^{r} \left( \exp (x_{i}\sum_{k=m+1}^{n}g_{k}) - 1
	\right)  < \epsilon \enspace. 
	\end{eqnarray}

	\noindent Now it is enough if we show that
	\begin{eqnarray}
	\label{Equation:WithoutSumEpsilonInequality}
	\exists N_{i} = N_{i}(\frac{\epsilon}{r}) \in {\dZ}^{+} \ni 
	n, m \geq N_{i} \Longrightarrow \\ \exp(
	x_{i}\sum_{k=m+1}^{n}g_{k}) - 1 \leq 
	\frac{\epsilon}{r}\enspace,  \:\:\: \forall i=1 \ldots r \enspace. \nonumber
	\end{eqnarray}

	\noindent For $N = \max\{N_{i}: i=1 \ldots r\}$

	\begin{equation}
	n, m \geq N \Rightarrow \exp( x_{i}\sum_{k=m+1}^{n}g_{k}) - 1 \leq
	\frac{\epsilon}{r} \enspace, \:\:\: \forall i=1 \ldots r \enspace ,
	\end{equation}

	\noindent which gives us

	\begin{equation}
	n, m \geq N \Rightarrow \sum_{i=1}^{r}\exp(
	x_{i}\sum_{k=m+1}^{n}g_{k}) - 1 \leq 
	\sum_{i=1}^{r}\frac{\epsilon}{r} = \epsilon \enspace.
	\end{equation}

	\noindent Now to
	assert~(\ref{Equation:WithoutSumEpsilonInequality}) it is 
	enough, for a fixed $x \in E_{\varphi}$,  if we have
	following

	\begin{eqnarray*}
	\forall \epsilon' > 0, 
	  \exists N'=N'(\epsilon') \in {\dZ}^{+} \ni 
	n, m \geq N' \Longrightarrow \\ \exp \left( x
	\sum_{k=m+1}^{n}g_{k} \right) - 1 
	  \leq \epsilon' \enspace. 
	\end{eqnarray*}

	\noindent Note that $\epsilon'$ can be chosen as $\epsilon' =
	\frac{\epsilon}{r}$, and $\epsilon'$ is arbitrary since
	$\epsilon$ arbitrary. Since
	\[
	\exp( x \sum_{k=m+1}^{n}g_{k}) - 1 \leq \epsilon' \Longrightarrow
  	\sum_{k=m+1}^{n}g_{k} \leq
	\frac{\ln \left(\epsilon' + 1 \right)}{x} \]

	\noindent it is enough if

	\begin{eqnarray}
	\label{Equation:CauchyOfS}
	\forall \epsilon'' > 0,
	  \exists N''=N''(\epsilon'') \in {\dZ}^{+} \ni 
	n, m \geq N'' \Longrightarrow \nonumber \\ \sum_{k=m+1}^{n}g_{k}
	\leq \epsilon'' \enspace.
	\end{eqnarray}

	\noindent Note that $\epsilon''$ can be chosen as $\epsilon'' = 
	\frac{\ln \left(\epsilon' + 1 \right)}{x}$ for a fixed $x \in
	E$ and $\epsilon''$ is arbitrary since
	$\epsilon'$ is arbitrary.

	\noindent Since $\epsilon''$ is
	arbitrary~(\ref{Equation:CauchyOfS}) can be asserted if the
	sequence 

	\[ \left\{\sum_{k=1}^{n} g_{k} \right\}_{(n)} \]
	is Cauchy by the definition of Cauchy sequence.

	\end{proof}

\section{Simulation Results}
\label{Section:SimulationResults}

  \subsection{Choice of $\{g_{k}\}$}

\begin{figure}[htbp]
  	\centering
        \fbox{
	\includegraphics[width = 0.45 \textwidth]
	{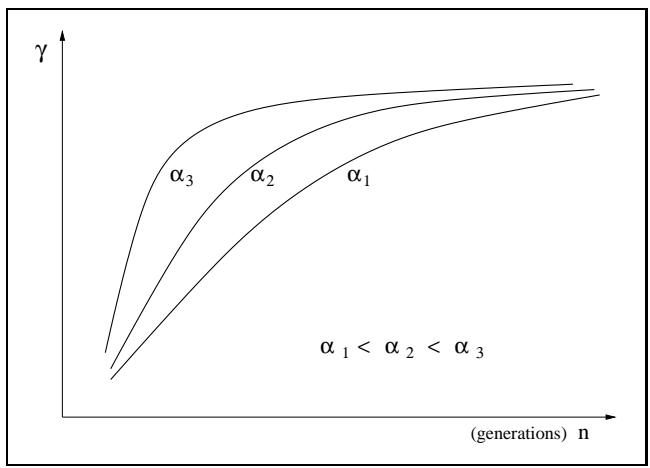}
        }
  	\caption{Cauchy Annealing Schedules for Different Values of
	$\alpha$
	where $\gamma$ is defined according
	to~(\ref{Equation:ChoiceOfAnnealingSchedule})
	}  
  	\label{Figure:CauchyAnneling_DifferentValuesOfAlpha}
	\end{figure}

	As a specific case, for $\{g_{k}\}$ to
	satisfy~(\ref{Equation:ChoiceOfGamma}),  we choose
	\begin{equation}
	g_{k} = g_{0}\frac{1}{k^{\alpha}},
	\end{equation}
	where $g_{0}$
	is any constant and $\alpha >1$. Since
	$\left \{ \sum_{k=1}^{n} \frac{1}{k^{\alpha}} \right \}_{n}$
	is a Cauchy sequence for any $\alpha >
	1$~\cite{Rudin:1964:PrinciplesOfMathematicalAnalysis},  
	$\left \{g_{0}\sum_{k=1}^{n}\frac{1}{k^{\alpha}} \right
	\}_{n}$ is  also a Cauchy sequence. In this specific choice of
	sequence, $\alpha$ plays an important role in the annealing
	schedule (see
	Figure~\ref{Figure:CauchyAnneling_DifferentValuesOfAlpha}). 

	Here we give simulation results using the annealing schedule
	$\{\gamma_{n}\}$ defined as
	\begin{equation}
	\label{Equation:ChoiceOfAnnealingSchedule}
	\gamma_{n} = \sum_{k=1}^{n} g_{k} = g_{0} \sum_{k=1}^{n} \frac{1}{k^{\alpha}}.
	\end{equation}

  \subsection{Results}

	We discuss the simulations conducted to study the annealing
	schedule for Boltzmann selection proposed in this paper. 
	We compare three selection
	mechanisms viz., proportionate selection ({\it proportionate}), Boltzmann selection
	with constant $\gamma$ ({\it Boltzmann}) and Boltzmann
	selection with proposed Cauchy
	annealing schedule $\{\gamma_{n}\}$ ({\it Cauchy-Boltzmann}). 
	We study multi-variable function optimization in the framework of
	genetic algorithms. Specifically, we use the
	following functions~\cite{Muhlenbein:1993:PredictiveModelsForBGA}:

	\begin{itemize}

	  \item Rastrigin's function:

		$f_{6}(\vec{x}) = lA + \sum_{i=1}^{l} { x_{i}^{2} - A
		\cos(2 \pi x_{i}) }$, \\ where $A=10$ ; $-5.12 \leq x_{i}
		\leq 5.12 $

	  \item Griewangk's function:

		$f_{8}(\vec{x}) = \sum_{i=1}^{l}
		\frac{{x_{i}}^{2}}{4000} -  \prod_{i=1}^{l} \cos( \frac{x_{i}}{\sqrt{i}}) +
		1$, \\ where $-600 \leq x_{i} \leq 600$

	  \item Ackley's function:

		$f_{9}(\vec{x}) = -20 \exp(-0.2 \sqrt{ \frac{1}{l}
		\sum_{i=1}^{l} {x_{i}}^{2}} ) \\ - \exp( \frac{1}{l}
		\sum_{i=1}^{l} \cos(2 \pi x_{i})) +
		20 + e$, \\where $-30 \leq x_{i} \leq 30$ 

	  \item Schwefel's function:

		$f_{7}(\vec{x}) = \sum_{i=1}^{l} - x_{i} \sin (
		\sqrt{\left| x_{i} \right|} )$, \\where $-500 \leq x_{i}
		\leq 500$

	\end{itemize}

\begin{figure}[htbp]
  	\centering
	\includegraphics[width = 0.45 \textwidth]
	{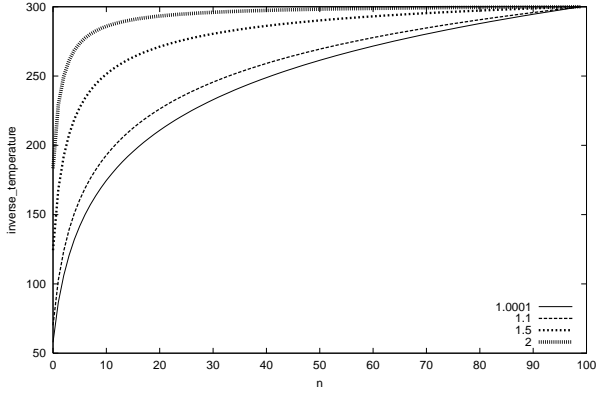}
  	\caption{ $\gamma_{n}$ for $\alpha = 1.0001, 1.1, 1.5, 2$
	where $\gamma_{n}$ is defined according
	to~(\ref{Equation:ChoiceOfAnnealingSchedule})
	}  
  	\label{Figure:Gamma_DifferentValuesOfAlpha}
	\end{figure}

	The following parameter values have been used in all the
	experiments: 
	\begin{itemize}

	  \item Each $x_{i}$ is encoded with 5
	  bits and $l=15$ i.e search space is of size $2^{75}$

          \item Population size $n_{P} = 150 $

	  \item For Boltzmann selection the inverse temperature $\gamma
	  = 300$. For Boltzmann selection with annealing, we vary
	  $\alpha = 1.0001, 1.1, 1.5, 2$ and we chose $g_{0}$ for each
	  value of $\alpha$ in such a way that, $\gamma_{100} = 300$
	  where 100 is the total number of generations for each
	  process. Figure~\ref{Figure:Gamma_DifferentValuesOfAlpha}
	  shows the plots of values of $\gamma_{n}$ for $\alpha
	  = 1.0001, 1.1, 1.5, 2$.

	  \item For all the experiments probability of uniform
	        crossover is $0.8$  
	        and probability of mutation is below $0.1$

	  \item Each simulation is performed $17$ times to get the
	  average behavior of the process  

	\end{itemize}

 	\begin{figure}[htbp]
   	\centering
 	\includegraphics[width = 0.5 \textwidth
   	]{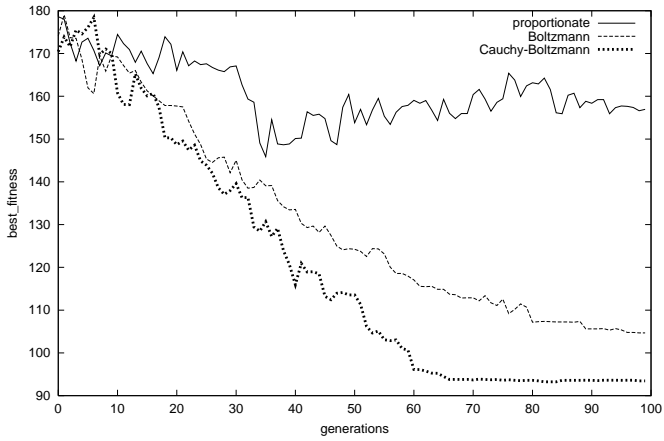}
   	\caption{Rastrigin: $\alpha = 2$ (Multiple Runs)}
   	\label{MultiPlot:R4}
 	\end{figure}

 	\begin{figure}[htbp]
   	\centering
 	\includegraphics[width = 0.5 \textwidth
   	]{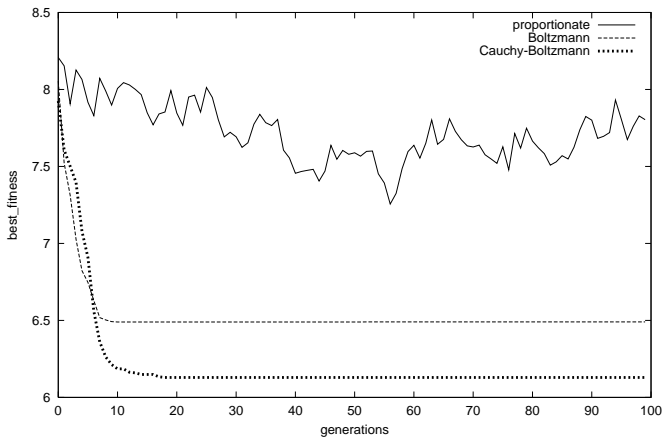}
   	\caption{Ackley: $\alpha = 1.1$ (Multiple Runs)}
   	\label{MultiPlot:A2}
 	\end{figure}

 	\begin{figure}[htbp]
   	\centering
 	\includegraphics[width = 0.5 \textwidth
   	]{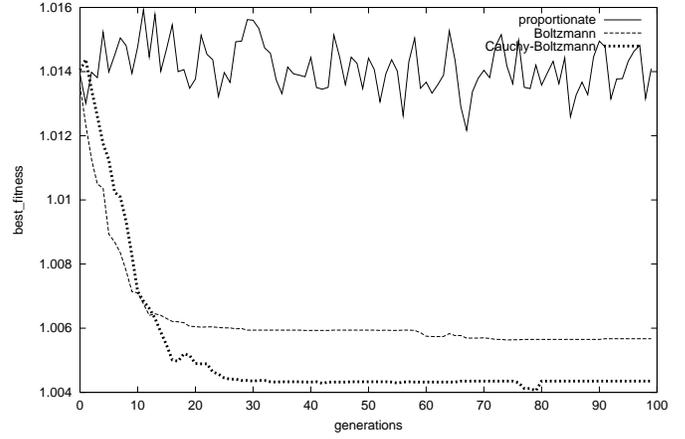}
   	\caption{Griewangk: $\alpha = 1.1$ (Multiple Runs)}
   	\label{MultiPlot:G2}
 	\end{figure}

 	\begin{figure}[htbp]
   	\centering
 	\includegraphics[width = 0.5 \textwidth
   	]{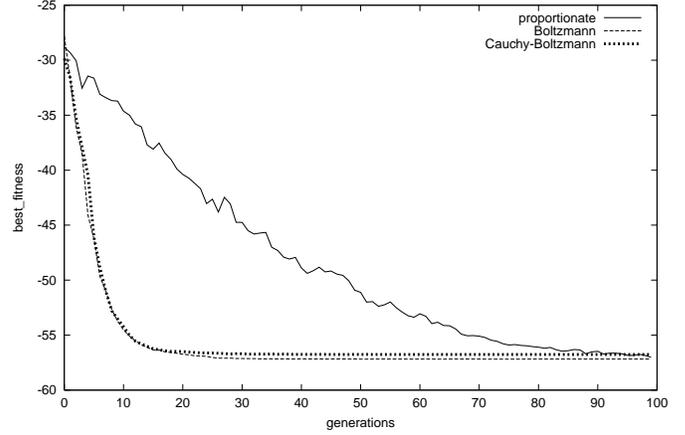}
   	\caption{Schwefel: $\alpha = 1.5$ (Multiple Runs)}
   	\label{MultiPlot:S3}
 	\end{figure}


 	\begin{figure}[htbp]
   	\centering
 	\includegraphics[width = 0.5 \textwidth
   	]{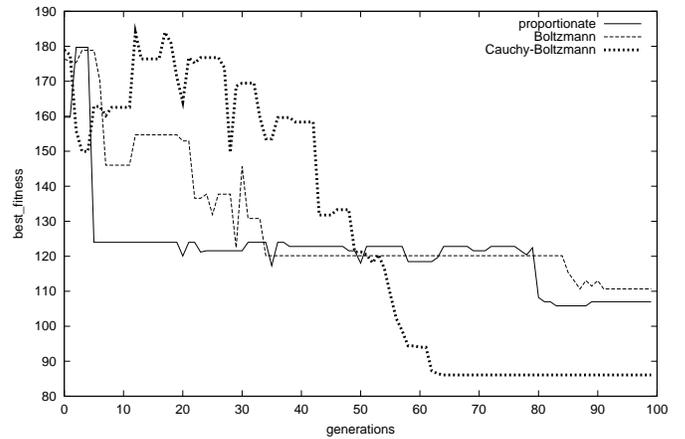}
   	\caption{Rastrigin: $\alpha = 2$ (Single Run)}
   	\label{SinglePlot:R4}
 	\end{figure}

 	\begin{figure}[htbp]
   	\centering
 	\includegraphics[width = 0.5 \textwidth
   	]{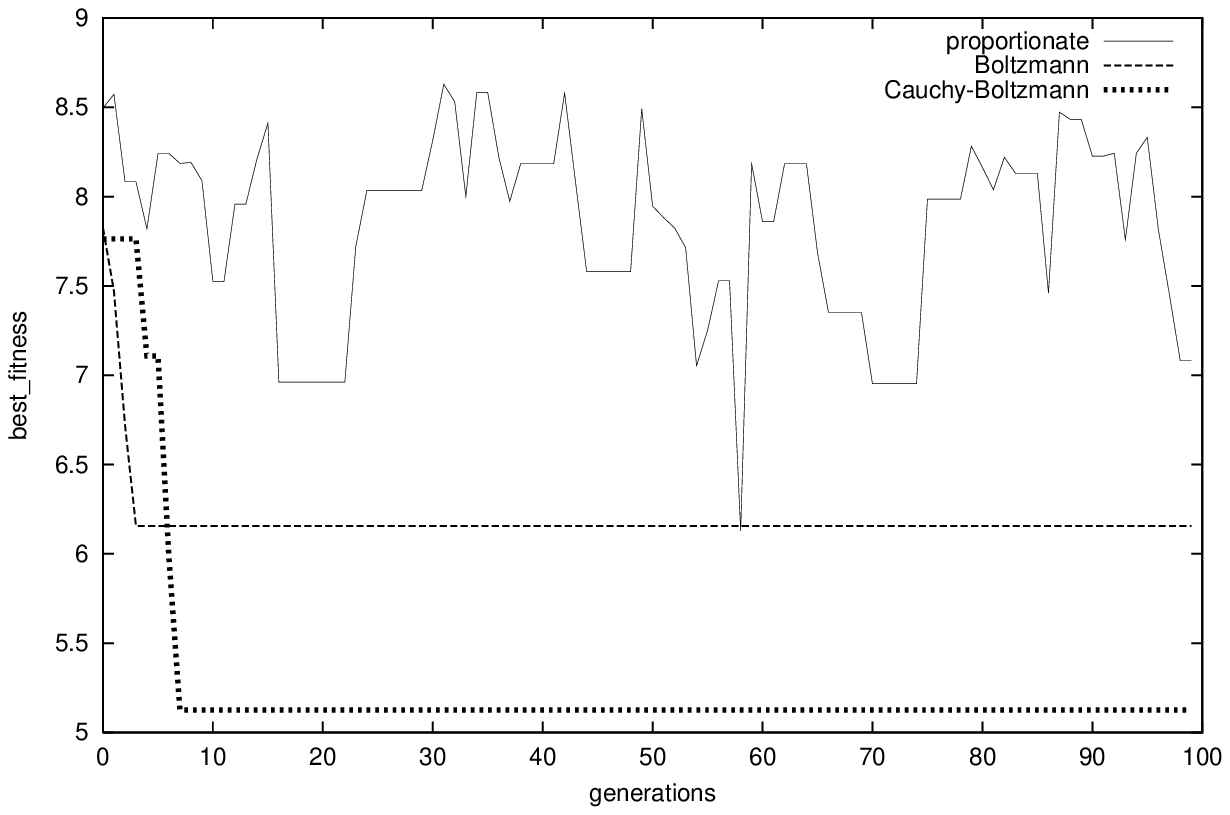}
   	\caption{Ackley: $\alpha = 1.1$ (Single Run)}
   	\label{SinglePlot:A2}
 	\end{figure}

	From various simulations we observed that when the problem
	size is small (for example smaller values of $l$) all the
	selection mechanisms perform equally well. Boltzmann selection
	is 
	effective when we increase the problem size. In the case of
	Boltzmann selection with constant $\gamma$, one has to increase
	the value of $\gamma$ when the problem size is large.
	Note that choice of
	parameter $\alpha$ is very important for Cauchy annealing
	schedule and it depends on the specific problem. Here we have
	given results corresponding to the best values of $\alpha$.
	Figures~\ref{MultiPlot:R4}, \ref{MultiPlot:A2}, \ref{MultiPlot:G2},
	\ref{MultiPlot:S3}, show the plots for behavior of the
	process when averaged over multiple
	runs. Figures~\ref{SinglePlot:R4} and~\ref{SinglePlot:A2} show
	plots for single run.
	Our simulations showed that Boltzmann selection with the
	Cauchy annealing schedule performs better than other
	mechanisms.

\section{Conclusions}
\label{Section:Conclusion}

	In this paper we derived an annealing schedule for inverse
	temperature parameter in the Boltzmann selection scheme, which
	is based
	on Cauchy criteria for Boltzmann selection schedule. Usage of
	Cauchy criteria for Boltzmann selection schedule is justified
	by the hypothesis: as process goes on

	\begin{itemize}
	  \item selection strength should increase,
	  \item difference between the selection strengths should decrease.
	\end{itemize}

	We have given alternative formalism for selection
	mechanisms based on the fitness distributions. We have also
	given a new measure for selection strength which is
	suitable for theoretical analysis.

	Using the above formalism, we presented an important mathematical
	result for Boltzmann selection schedule; using which we derived
	the annealing schedule. Cauchy annealing schedule is a generalized
	mechanism from which one can choose different specific
	sequences for annealing based on the problem at hand. 

	Our simulation results justify the hypothesis we
	presented and the utility of techniques we used; they also
	support usage of the mathematical
	results we presented, in practice. We conducted experiments
	using specific annealing schedule, where one can choose the
	speed of (inverse) annealing. We compared our results
	with 
	algorithms with proportionate selection, Boltzmann selection
	without annealing schedule and Boltzmann selection with
	the proposed annealing schedule. We found that with an
	appropriate choice of speed of annealing, algorithms with
	annealing schedule outperform other methods.

	This analysis does not consider any of the genetic
	operators. Our future work would involve comprehensive
	analysis which leads to more generalized selection schedules
	based on the techniques presented in this paper.

	One important consequence of techniques we developed in this
	paper would be proving convergence of the process. If one can
	show that the underlying space, for example spaces of NFDs, is
	{\em complete} (see Appendix for the definition of complete
	metric 
	space), one can conclude the convergence of evolutionary 
	process, based on the {\em Cauchy criteria}.

\begin{acknowledgments}
We would like to thank Dr. Gary Fogel for valuable suggestions on
simulations. Research work reported here is supported in part by AOARD
Grant F62562-03-P-0318.   
\end{acknowledgments}

\appendix

\section{Metric Spaces}
 	Here we present some basic concepts of metric spaces used in
 	this paper.

 	Let $X$ be any set. A function $d: X \times X \rightarrow
 	\bbbr$ is said to be metric on $X$ if
 	\begin{enumerate}

 	  \item $d(x,y) \geq 0 $ and $d(x,y) = 0 \Leftrightarrow x =
 	y \enspace, \:\:\: \forall x,y \in X$

 	  \item $d(x,y) = d(y,x) \enspace, \:\:\: \forall x,y \in X$
		
 	  \item $d(x,y) \leq d(x,z) + d(z,y) \enspace,\:\:\: \forall x,y,z \in
 	  X$ (Triangular inequality)  

 	\end{enumerate}
 	Example of metric space is $\bbbr$ with $\mid . \mid$ as a
 	metric.

 	A sequence $\{x_{n}\}$ is said to be Cauchy sequence if 
 	\[ \forall \epsilon > 0 , \exists N = N(\epsilon) \in
 	{\dZ}^{+} \ni \]
 	\[  n, m \geq N \Rightarrow d(x_{n},x_{m}) < \epsilon
 	\enspace.\]

 	We say metric space $(X,d)$ is complete if every Cauchy
 	sequence in $X$ converges.

\bibliographystyle{apsrmp}
\bibliography{papi}

\end{document}